\documentclass[runningheads,a4paper]{llncs}
\usepackage{amssymb}
\usepackage{graphicx,epsfig,setspace,subfig,url,amsmath}
\usepackage{algorithm}
\usepackage[noend]{algpseudocode}
\usepackage{longtable}
\usepackage{array}
\usepackage{amsmath}
\usepackage{mathtools}
\usepackage{caption}
\usepackage{multirow}
\usepackage{relsize}
\usepackage{color}

\usepackage[T1]{fontenc}
\usepackage[font=small,labelfont=bf,tableposition=top]{caption}

\newcommand{\keywords}[1]{\par\addvspace\baselineskip\noindent\keywordname\enspace\ignorespaces#1}
 \normalsize

\begin{document}

\title{Unsupervised Task Design to Meta-Train \\ Medical Image Classifiers \thanks{Supported by Australian Research Council through grant DP180103232.}}
\institute {$^{\dagger}$Australian Institute for Machine Learning, The University of  Adelaide  \\ $^{\ddagger\ddagger}$Institute for Systems and Robotics, Instituto Superior Tecnico, Portugal}
\author{Gabriel Maicas$^{\dagger}$ \qquad  Cuong Nguyen$^{\dagger}$ \qquad Farbod Motlagh$^{\dagger}$ \\ Jacinto C. Nascimento$^{\ddagger\ddagger}$ \qquad Gustavo Carneiro$^{\dagger}$}

\maketitle

\vspace{-.1in} 
\begin{abstract}

Meta-training has been empirically demonstrated to be the most effective pre-training method for few-shot learning of medical image classifiers (i.e., classifiers modeled with small training sets).  
However, the effectiveness of meta-training relies on the availability of a reasonable number of hand-designed classification tasks, which are costly to obtain, and consequently rarely available.
In this paper, we propose a new method to unsupervisedly design a large number of classification tasks to meta-train medical image classifiers.
We evaluate our method on a breast dynamically contrast enhanced magnetic resonance imaging (DCE-MRI) data set that has been used to benchmark few-shot training methods of medical image classifiers. 
Our results show that the proposed unsupervised task design to meta-train medical image classifiers builds a pre-trained model that, after fine-tuning, produces better classification results than other unsupervised and supervised pre-training methods, and competitive results with respect to meta-training that relies on hand-designed classification tasks.

\vspace{-.2in} 
-\keywords{meta-training, unsupervised learning, unsupervised task design, breast image analysis, magnetic resonance imaging, few-shot, pre-training, clustering.}
\end{abstract}

\section{Introduction}
\label{sec:intro}

The accuracy and robustness of deep learning based medical image classifiers is generally positively correlated with the size of the annotated training set used during the modelling process~\cite{litjens2017survey}.
However, large annotated training sets are expensive and not readily available for some medical image analysis applications, such as breast screening from DCE-MRI~\cite{mcclymont2014fully}. Therefore, training medical image classifiers with small annotated training sets has become a highly investigated topic, particularly after the advent of deep learning~\cite{litjens2017survey}.

\begin{figure}[t]
\begin{center}
\begin{tabular}{c}
\includegraphics[width=0.98\textwidth]{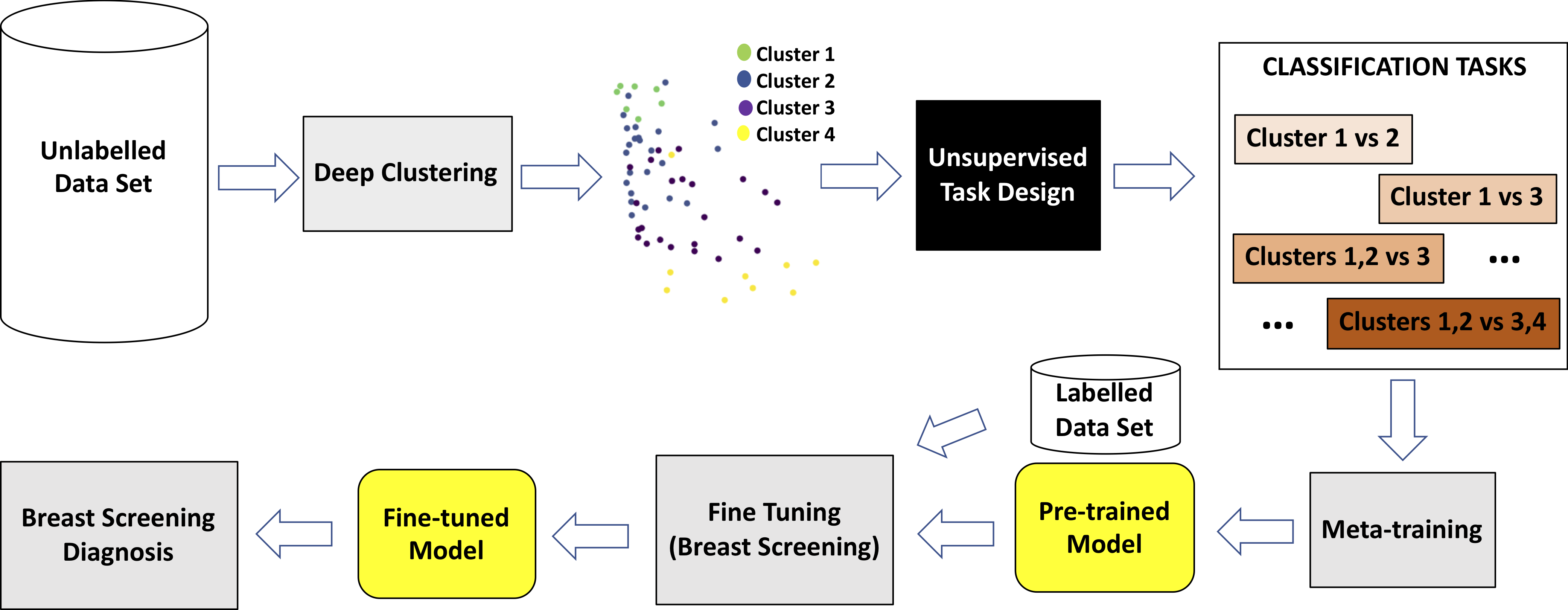} \\
\end{tabular}
\end{center}
\caption{Unsupervised task design to meta-train medical image classifiers.  Deep clustering~\cite{caron2018deep}  produces a set of clusters that are used in the unsupervised design of classification tasks.  These tasks are used in a meta-training process to produce a pre-trained model that can be fine-tuned to new classification tasks using small labelled training sets, in this paper represented by the breast screening problem from DCE-MRI~\cite{maicas2018training}.}
\label{fig:intro}
\end{figure}

The most competitive medical image classifiers are currently based on convolutional neural networks (CNNs)~\cite{litjens2017survey} that need large training sets to be properly modelled. 
To reduce the need for such large annotated sets, pre-training approaches have been explored in medical image analysis, where the most relevant for our paper are: 1) supervised pre-training using independent data sets~\cite{bar2015chest}, where the model is pre-trained by solving a classification problem in a different data set; 2) unsupervised pre-training using clustering~\cite{caron2018deep}, where the model is pre-trained by performing clustering without any knowledge about the ground truth labels; and 3) unsupervised pre-training using input reconstruction~\cite{dong2018learning}, where the model is pre-trained by reconstructing the input images of the training set.  
Arguably, the main issue with these pre-training methods is that their objective functions are irrelevant for the medical image classifier being developed downstream.  
Alternatively, the need for pre-training methods can be alleviated with the use of other types of training methods, such as multiple instance learning (MIL)~\cite{zhu2017deep} or multi-task learning~\cite{xue2018full}, but both methods still need large training sets.  More recently, a pre-trained model produced by supervised meta-training (i.e., a meta-training process that depends on hand-designed classification tasks) showed superior performance compared to the previously described pre-training methods~\cite{maicas2018training}. Nevertheless, these promising meta-training results are counterbalanced by an unappealing need of an expensive hand-designing process to produce the classification tasks~\cite{maicas2018training}.  Given the high cost of this process, the availability of a large number of hand-designed classification tasks is rare, which hampers the exploration of meta-training for medical image classifiers.

In this paper, we propose a new method to unsupervisedly produce a large number classification tasks to meta-train medical image classifiers. To this end, we use deep clustering~\cite{caron2018deep} to automatically build image clusters that can be grouped in different ways to enable the design of multiple classification tasks employed in the meta-training process -- see Fig.~\ref{fig:intro}. 
We evaluate our method on the breast screening classification task from a breast DCE-MRI data set that has been used to benchmark few-shot training algorithms of medical image classifiers~\cite{maicas2018training}. Results show that our proposed approach produces classification results that are significantly better than other unsupervised and supervised pre-training methods, and competitive to supervised meta-training.

\section{Literature Review}
\label{sec:lit_review}
 
DCE-MRI is a recommended image modality in breast screening programs for patients at high-risk~\cite{mainiero2017acr}. However, DCE-MRI interpretation is time-consuming and prone to high inter-observer variability~\cite{grimm2015interobserver}. Thus, computer-aided diagnosis (CAD) systems are being developed to assist radiologists increase their diagnosis sensitivity~\cite{Vreemann2018} and specificity~\cite{meinel2007breast}, and reduce analysis time. However, the development of CAD systems for breast DCE-MRI is challenging due in part to the small size of annotated data sets available for training.

Meta-training has been shown to be an effective strategy to improve the learning of classifiers using relatively small training sets~\cite{finn2017model}. For instance, Maicas \textit{et al.}~\cite{maicas2018training} proposed the use of hand-designed breast classification tasks to meta-train a model that was then fine-tuned to solve the breast screening task. Results showed that this method improves over other strategies to train classifiers from small data sets, such as MIL~\cite{zhu2017deep} and multi-task learning~\cite{xue2018full}. However, the method proposed in~\cite{maicas2018training} relies on costly hand-designed classification tasks.

Similarly to our paper, Hsu \textit{et al.}~\cite{hsu2018unsupervised} proposed an unsupervised method to design computer vision classification tasks for meta-training. Results showed that this approach produced worse classification performance than meta-training modelled with hand-designed tasks (i.e., supervised meta-training). We believe that the reason behind this drop in performance lies in the large number of hand-designed tasks already available for supervised meta-training in computer vision applications~\cite{hsu2018unsupervised}, enabling a good classification performance baseline. The difficulty to obtain a large number of hand-designed tasks for medical image classification problems means that the number of these hand-designed tasks will be small, which may result in a relatively low classification performance baseline. We hypothesize that our proposed method that unsupervisedly designs a large number of classification tasks to meta-train a medical image classifier can achieve a classification performance that is at least comparable to supervised meta-training~\cite{maicas2018training} trained with a small number of hand-designed tasks. Our proposed method has the advantage that it does not rely on costly hand-designed tasks.


\section{Data Set and Methods}
\label{sec:methodology}

\subsection{Data set}
\label{sec:dataset}

The data set is represented by ${\cal D} = \left \{ \left ( \mathbf{v}_i, \mathbf{t}_i, b_i, y_i \right ) \right \}_{i=1}^{|{\cal D}|}$, where $\mathbf{v}:\Omega \rightarrow \mathbb R$ corresponds to the first DCE-MRI subtraction volume 
($\Omega$ denotes the volume lattice)~\cite{gilbert2018personalised}, 
$\mathbf{t}:\Omega \rightarrow \mathbb R$  represents the T1-weighted MRI only used to separate the left and breast regions of the volume, $b \in \{ \text{left},\text{right} \}$ indicates the left or right breast, and $y \in \mathcal{Y} = \{0,1\}$ indicates the classification label: no malignant findings, or malignant findings, respectively.

\subsection{Deep Clustering to Unsupervisedly Design Classification Tasks}
\label{sec:unsupervised_tasks}

The proposed unsupervised task design method builds several binary classification tasks from image groups formed by deep clustering~\cite{caron2018deep}. The training of deep clustering alternates an optimisation of two objective functions~\cite{caron2018deep}. We denote the $\theta$-parameterised model that produces the unsupervised learning features by $f_{\theta}(\mathbf{v}) \in \mathbb R^D$ and the $\omega$-parameterised classifier that produces a pseudo-label representing one of the unknown $K$ classes and is placed on top of $f_{\theta}(.)$ by $g_{\omega}(f_{\theta}(\mathbf{v})) \in \{0,1\}^K$. The first objective function is the cross-entropy loss $\ell(.)$ with respect to the pseudo-labels $\{\widetilde{\mathbf{y}}_i\}_{i=1}^{|\mathcal{D}|}$, with $\widetilde{\mathbf{y}} \in \widetilde{\mathcal{Y}}=\{0,1\}^K$,

\begin{equation}
    \min_{\theta,\omega} \frac{1}{|\mathcal{D}|}\sum_{i=1}^{|\mathcal{D}|} \ell(g_\omega(f_{\theta}(\mathbf{v}_i)) , \widetilde{\mathbf{y}}_i),
\label{eq:kmeans_obj1}
\end{equation}
which is used to estimate the optimal $\theta^*$ and $\omega^*$.  The second objective function finds the $K$ centroids, denoted by $\mathbf{C} \in \mathbb R^{D \times K}$, and pseudo-labels $\widetilde{\mathbf{y}}$ with
\begin{equation}
    \min_{\mathbf{C}} \frac{1}{|\mathcal{D}|}\sum_{i=1}^{|\mathcal{D}|}\min_{\widetilde{\mathbf{y}}_i} \Vert f_{\theta}(\mathbf{v}_i) - \mathbf{C}\widetilde{\mathbf{y}}_i  \Vert_{2}^{2},
\label{eq:kmeans_obj2}
\end{equation}
where $\widetilde{\mathbf{y}}_i$ is a $K$-dim one-hot vector.

Each step of the optimization above will generate new values for the model parameters, centroids and pseudo-labels.  We extend deep clustering~\cite{caron2018deep} with a model selection process based on maximising the Silhouette coefficient that measures clustering quality~\cite{rousseeuw1987silhouettes} with
\begin{equation}
    \kappa = \frac{1}{|\mathcal{D}|}\sum_{i=1}^{|\mathcal{D}|}\frac{b(i) - a(i)}{\max \left ( a(i),b(i) \right ) },
    \label{eq:silhoutte}
\end{equation}
where $a(i)$ represents the average $\ell_2$ distance between $f_{\theta}(\mathbf{v}_i)$ and all points $f_{\theta}(\mathbf{v}_j)$ where $i \neq j$ and $\widetilde{\mathbf{y}}_i = \widetilde{\mathbf{y}}_j$; and $b(i)$ denotes the smallest average $\ell_2$ distance between $f_{\theta}(\mathbf{v}_i)$ and $f_{\theta}(\mathbf{v}_j)$ where $i \neq j$ and $\widetilde{\mathbf{y}}_i \neq \widetilde{\mathbf{y}}_j$.  

The unsupervised design of classification tasks is based on the formation of $L$ binary classification problems derived from the pseudo-labels obtained from~\eqref{eq:kmeans_obj2}. Each of these $L$ binary classification problems is built by randomly selecting 2 nonempty and disjoint subsets \(\mathcal{K}_{l}^{(0)}\) and \(\mathcal{K}_{l}^{(1)}\) from the pseudo label set \(\{1, 2, \dots, K\}\) and labelling their corresponding data points as class 0 and 1, respectively. 
Note that the number of classification tasks for a given  $K$ is $L=\sum_{i=1}^{n-1}\sum_{k=1}^{\min(i,n-i)}\frac{\binom{n}{i}\times\binom{n-i}{k}}{1+\delta(i-k)}$, where $\binom{A}{B}$ denotes the binomial coefficient, and $\delta(.)$ represents the Dirac delta function.

\subsection{Meta-training with the Unsupervised Classification Tasks}
\label{sec:meta_training}

Meta-training estimates the parameters of a meta-learner, so it can be used as a pre-trained model that is efficiently fine-tuned to previously unseen classification tasks, using small annotated training sets~\cite{finn2017model}.  The algorithm assumes that there exists a task distribution $\mathcal{T}$, from which each classification task $\mathcal{T}_l$ is drawn, where each task comprises a training set $\{ \mathbf{v}_i^{(l,t)}, \widetilde{y}_i^{(l,t)} \}_{i=1}^{M}$ and a testing set $\{ \mathbf{v}_i^{(l,v)}, \widetilde{y}_i^{(l,v)} \}_{i=1}^{N}$, with $M << N$ and $M+N = |\mathcal{T}_l|$.  Meta-training iteratively samples $T$ tasks from $\mathcal{T}$, and re-trains a multi-target classifier for those tasks using the training and testing sets defined above.

We use the MAML meta-training~\cite{grant2018recasting} that consists of a Bayesian hierarchical model, where $\psi$ denotes the classifier meta parameter, and $\phi_l$ represents the parameter for task $\mathcal{T}_l$.  The meta-training objective function is defined by:
\begin{equation}
    \begin{aligned}[b]
    \max_{\psi} \log p(  \mathcal{Y}^{(v)}_{l=1..T} \vert \mathcal{Y}^{(t)}_{l=1..T}, \mathcal{V}^{(v)}_{l=1..T},  \mathcal{V}^{(t)}_{l=1..T}, \psi),
    \end{aligned}
    \label{eq:meta_learning_objective}
\end{equation}
where $T$ is the number of tasks per meta-training iteration, $\mathcal{Y}^{(v)}_l = \{ \widetilde{y}_i^{(l,v)} \}_{i=1}^{N}$,
$\mathcal{Y}^{(t)}_l = \{ \widetilde{y}_i^{(l,t)} \}_{i=1}^{M}$,
$\mathcal{V}^{(v)}_l = \{ \mathbf{v}_i^{(l,v)} \}_{i=1}^{N}$, and
$\mathcal{V}^{(t)}_l = \{ \mathbf{v}_i^{(l,t)} \}_{i=1}^{M}$.  In \eqref{eq:meta_learning_objective}, we have
\begin{equation}
\begin{aligned}[b]
    \log p(  \mathcal{Y}^{(v)}_{l=1..T} \vert \mathcal{Y}^{(t)}_{l=1..T}, \mathcal{V}^{(v)}_{l=1..T},  \mathcal{V}^{(t)}_{l=1..T}, \psi) 
    & \ge   \sum_{l=1}^{T} \mathbb{E}_{p(\phi_{l} \vert \mathcal{Y}_{l}^{(t)}, \mathcal{V}^{(t)}_l, \psi)} \left[  \log p(\mathcal{Y}_{l}^{(v)} \vert \mathcal{V}^{(v)}_l, \phi_{l}) \right],
\end{aligned}
\label{eq:maml_likelihood}
\end{equation}
where the lower bound is derived from Jensen's inequality~\cite{bishop2006pattern}.  Therefore, the maximisation in~\eqref{eq:meta_learning_objective} is approximated with the lower bound maximisation in~\eqref{eq:maml_likelihood}, where the posterior $p(\phi_{l} \vert \mathcal{Y}_{l}^{(t)}, \mathcal{V}^{(t)}_l, \psi)$ is approximated with a Dirac delta function at a local optimal task-specific model parameter $\phi_l^{*}$, with $p(\phi_{l} \vert \mathcal{Y}_{l}^{(t)}, \mathcal{V}^{(t)}_l, \psi) = \delta(\phi_l - \phi_l^{*})$. The local optimal model parameter $\phi_i^{*}$ is obtained with truncated gradient descent initialised by the meta parameters $\psi$:
\begin{equation}
    \phi_{l}^{*} = \psi - \alpha \nabla_{\phi_{l}} \left[-\log p(\mathcal{Y}_l^{(t)} \vert , \mathcal{V}^{(t)}_l, \phi_l)\right],
    \label{eq:point_estimate_value}
\end{equation}
where $\alpha$ is the learning rate, and the truncated gradient descent consists of a single step of \eqref{eq:point_estimate_value}. Maximising the lower bound of the log likelihood in \eqref{eq:maml_likelihood} represents the MAML algorithm in~\cite{finn2017model}, which produces a pre-trained model that can quickly learn new tasks drawn from $\mathcal{T}$.

\section{Experiments and Results}
\label{sec:Experiments} 


\subsection{Experimental Set-Up}

We evaluate our proposed method on a breast DCE-MRI data set~\cite{mcclymont2014fully} (formally defined in Sec.~\ref{sec:dataset}), which has previously been used to evaluate few-shot training methods~\cite{maicas2018training}. To allow a fair comparison with previous papers, we split the data set in a patient-wise manner into the same 
training, validation and testing sets, containing 45, 13, and 59 patients, respectively. We use the T1-weighted MRI to automatically extract the left and right breast regions from the first DCE-MRI subtraction volume~\cite{maicas2018training}. Each breast region is resized into a volume of $100\times100\times50$~\cite{maicas2018training}.
For the breast screening problem, only breasts that contain a malignant finding(s) are considered positive, while breasts with only benign findings or no findings are considered negative. There are 30, 9, and 38 positive and 60, 17, and 80 negative breasts in the training, validation and testing sets, respectively. 

The model $f_{\theta}(\bf{v})$ that unsupervisedly produces the volume features is a 3D Densenet~\cite{huang2017densely} composed of five dense blocks, each containing two dense layers. The features 
represent the input to the deep clustering algorithm, explained in Sec.\ref{sec:unsupervised_tasks}, with the number of clusters $K \in \{3,4,5\}$. 
The model that is meta-trained, and fine-tuned, has the same architecture as $f_{\theta}(.)$. During meta-training, we use a meta learning rate $\alpha=0.001$ in \eqref{eq:point_estimate_value}. 
At each meta-iteration, a meta-batch size of $T=4$ classification tasks is sampled according to a random or a curriculum learning strategy~\cite{maicas2018training}. The meta-trained model is fine-tuned to the breast screening task using the entire training set, where model selection is performed using the validation set and results are reported in the test set. 

The evaluation for the breast screening problem is based on the area under the ROC curve (AUC). We also measure the standard error utilising an estimate based on the Wilcoxon test~\cite{bradley1997use} that estimates confidence intervals based on the testing set. 
In this evaluation, we study the type of task sampling for meta-training, i.e. random, or curriculum learning~\cite{maicas2018training}, and the influence of the number of clusters $K$ in~\eqref{eq:kmeans_obj1}, used to build the tasks. 
We compare our method (U-MT) with the previously proposed supervised meta-training for the case where the breast screening task is included (S-MT (S)) and not included (S-MT (NS)) in the meta-training process. We also compare our method with: a) Densenet trained from scratch on the breast screening task; b) Densenet from (a) fine-tuned with MIL~\cite{zhu2017deep}; c) Densenet trained with multi-tasking (using hand-designed tasks)~\cite{maicas2018training}; d) Densenet pre-trained as a variational autocoder (i.e., unsupervised training) and fine-tuned for the breast screening task; and e) Densenet pre-trained with deep clustering (i.e., unsupervised training) and fine-tuned for the breast screening task.  All Densenet models of these competing methods have the same architecture as the meta-trained model described above.
The rationale for baselines (d) and (e) is to evaluate the effect of pre-training based on a reconstruction or a clustering scheme. With this purpose, we present results based on nearest neighbor classification and the fine-tuned classification model.

\subsection{Results}

\begin{figure}[t]
\begin{center}
\begin{tabular}{cccc}
\subfloat[]{\includegraphics[width=0.22\textwidth]{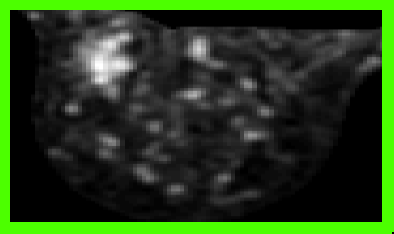}\label{img1}} &
\hspace{0.1in}
\subfloat[]{\includegraphics[width=0.22\textwidth]{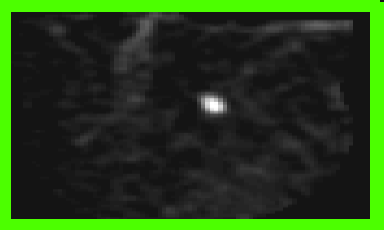}\label{img2}} &
\hspace{0.1in}
\subfloat[]{\includegraphics[width=0.22\textwidth]{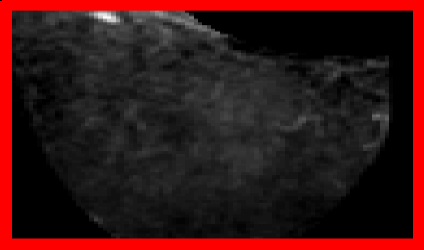}\label{img3}} &
\hspace{0.1in}
\subfloat[]{\includegraphics[width=0.22\textwidth]{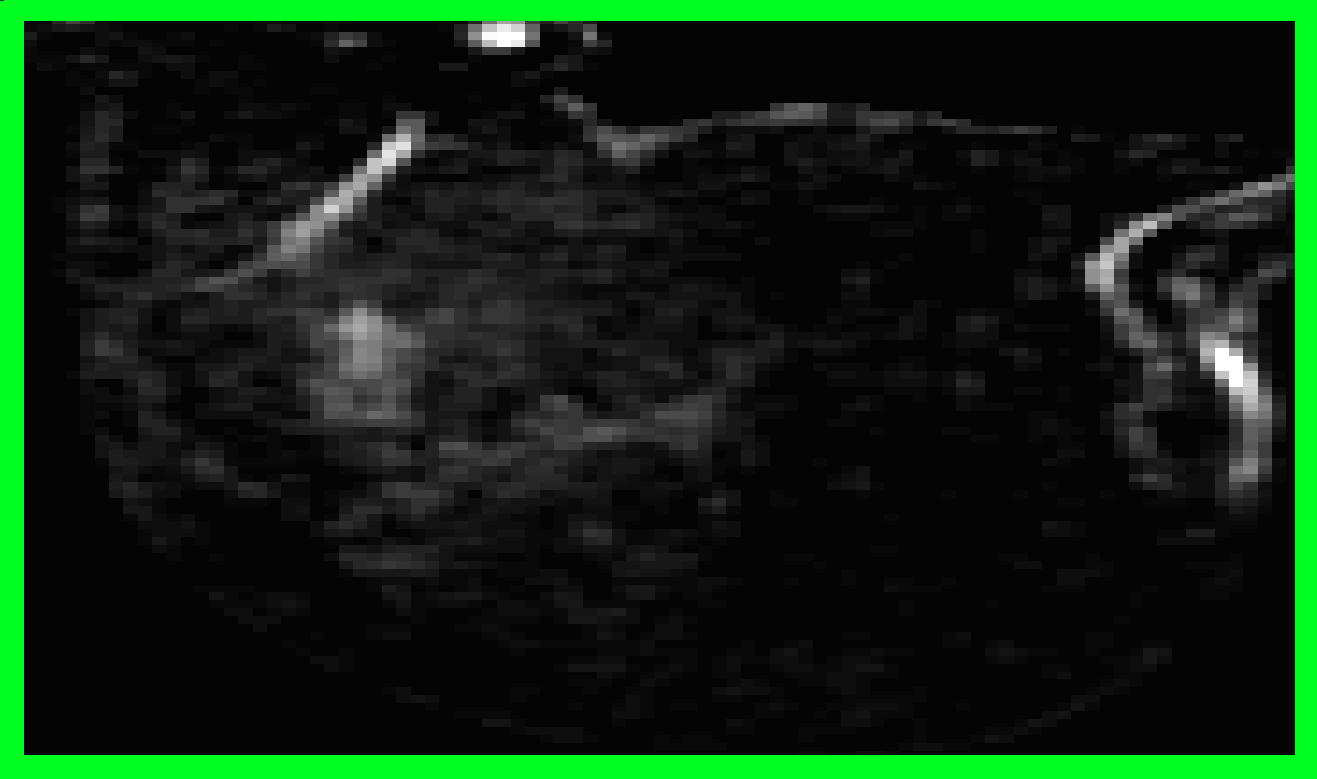}\label{img4}}
\end{tabular}
\end{center}
\caption{Example of breast screening diagnosis produced by our approach. Image (\ref{img1}) shows the correct positive diagnosis of a breast containing a malignant tumour. Image (\ref{img2}) shows the correct negative diagnosis of a breast with a benign tumour. Image (\ref{img3}) shows the incorrect positive classification of a breast containing no tumours. Image (\ref{img4}) shows the correct negative diagnosis of a breast with a benign tumour.}
\label{fig:breastExample}
\end{figure}

We show the AUC results ($\pm$ standard error) for breast screening baselines in Tab.~\ref{tab:Baselines}. Table~\ref{tab:unsupervisedRes} presents the results of meta-training, as a function of $K\in\{3,4,5\}$, with supervised and unsupervised task design using random and curriculum learning task sampling methods. Figure~\ref{fig:breastExample} presents examples of breast screening classification.

\begin{table}[]
\small
\centering
\begin{tabular}{|l|c|}
\hline
\textbf{Training Method Baseline} & \textbf{AUC}                                  \\
\hline
\hline
From Scratch~\cite{huang2017densely}          & $0.83 \pm 0.04$                                             \\
\hline
MIL based fine-tuning~\cite{zhu2017deep}               & $0.85 \pm 0.04$                                      \\
\hline
Multi-Task ~\cite{xue2018full}         & $0.85 \pm 0.04$                    \\
\hline
Variational Autoencoder + Nearest Neighbour       &         $0.61 \pm 0.06$            \\
\hline
Variational Autoencoder +  Fine-Tune in breast screening         & $0.84 \pm 0.04$                  \\
\hline
Deep Clustering + Nearest Neighbour           &       $0.53 \pm 0.06$               \\
\hline
Deep Clustering + Fine-Tune in breast screening          & $0.80 \pm 0.05$                     \\
\hline
\end{tabular}
\caption{AUC results ($\pm$ standard error) for breast screening baselines.}
\label{tab:Baselines}
\end{table}

\begin{table}[]
\small
\centering
\begin{tabular}{l|c|c|c|c|c|c|}
\cline{2-7}
                                                                                         & \multicolumn{3}{c|}{\textbf{Random}}                                      & \multicolumn{3}{c|}{\textbf{Curriculum}        }                                   \\ \cline{2-7} 
                                                                                         & \textbf{$K=3$} & \multicolumn{1}{c|}{\textbf{$K =4$}} & \textbf{$K=5$} & \textbf{$K=3$} & \multicolumn{1}{c|}{\textbf{$K=4$}} & \textbf{$K=5$} \\ \hline
\multicolumn{1}{|l|}{S-MT~\cite{maicas2018training} (S)}     & $0.86 \pm 0.04$                       & N/A                    & N/A                         &  $0.90\pm0.04$                         & N/A                    & N/A                         \\ \hline
\multicolumn{1}{|l|}{S-MT~\cite{maicas2018training} (NS)} & $0.85 \pm 0.04$                        & N/A                    & N/A                         & $0.89 \pm 0.04$                       & N/A                    & N/A                         \\ \hline
\multicolumn{1}{|l|}{\textbf{U-MT (Ours)}}                                            & $0.81 \pm 0.05$                        & $0.88 \pm 0.04$                   & $0.89 \pm 0.04$                      & $0.87 \pm 0.04$                       & $0.86 \pm 0.04$                   &         $0.88 \pm 0.04$                 \\ \hline
\end{tabular}
\caption{AUC for the breast screening task for our proposed method (U-MT)  as a function of the number of image clusters $K$ and the task sampling method (random and curriculum). We also present the results of supervised meta-training~\cite{maicas2018training} (S-MT) for the cases where the breast screening is included (labelled as S) and not included (labelled as NS) in the meta-training tasks. N/A indicates that the experiment is not feasible due to the lack of extra ground truth labels.}
\label{tab:unsupervisedRes}
 \end{table}

We measure the statistical significance of the difference in performance between our best performing approaches (Random with $K=5$ and Curriculum with $K=5$) and all baseline methods, obtaining a p-value $p \leq 0.001$ for all cases (unpaired two-tailed t-test).  Also, comparing our newly proposed U-MT (Random with $K=5$) and S-MT (S) (Curriculum with $K=3$)~\cite{maicas2018training}, we obtain a p-value $p > 0.05$.

 \section{Discussion and Conclusion}
\label{sec:dic}

We have presented a new method that unsupervisedly designs classification tasks to meta-train medical image classifiers. Our method significantly outperforms several baselines consisting of traditional pre-training methods based on variational autoencoder, deep clustering, MIL, and multi-task learning (see Tab.~\ref{tab:Baselines}). 
Our method also produces results comparable to the state-of-the-art set by meta-training using hand-designed tasks~\cite{maicas2018training} (see Tab.~\ref{tab:unsupervisedRes}). However, instead of using manually defined labels during meta-training, we unsupervisedly build classification tasks, allowing us to build a larger set of tasks, compared to the hand-designed ones. 
Also from Tab.~\ref{tab:unsupervisedRes}, we notice that larger number of tasks, which increases with the number of clusters (Sec.~\ref{sec:unsupervised_tasks}), generally implies better AUC results. This confirms our initial hypothesis that, differently from computer vision problems, automatically building tasks is of great importance for medical image classification problems, where image labels that allow a large number of tasks are costly to obtain. 
We also observe that sampling tasks according to curriculum learning provides a good improvement of accuracy compared to random task sampling for a small number of clusters ($K=3$), but not for larger number of tasks ($K=5$). We hypothesize that meta-training with curriculum learning sampling needs a larger number of meta-iterations to learn a curriculum that is better than random task sampling. Given the large number of tasks for $K \in \{4,5\}$, the meta-training process converged before the curriculum learning algorithm -- that deserves further research. 

\bibliographystyle{splncs}
\bibliography{biblio}
\end{document}